\documentclass[conference]{IEEEtran}
\IEEEoverridecommandlockouts
\usepackage{cite}
\usepackage{amsmath,amssymb,amsfonts}
\usepackage{algorithmic}
\usepackage{graphicx}
\usepackage{textcomp}
\usepackage{xcolor}
\def\BibTeX{{\rm B\kern-.05em{\sc i\kern-.025em b}\kern-.08em
    T\kern-.1667em\lower.7ex\hbox{E}\kern-.125emX}}
\begin{document}

\title{Classification of Alzheimer's Dementia vs. Healthy subjects by studying structural disparities in fMRI Time-Series of DMN\\}

\author{\IEEEauthorblockN{Sneha Noble, Chakka Sai Pradeep, Neelam Sinha, Thomas Gregor Issac}
\IEEEauthorblockA{\textit{Centre for Brain Research, IISc} \\
Bangalore, India \\
noblesneha98@gmail.com} 
}

\maketitle

\begin{abstract}
Time series from different regions of interest (ROI) of default mode network (DMN) from Functional Magnetic Resonance Imaging (fMRI) can reveal significant differences between healthy and unhealthy people. Here, we propose the utility of an existing metric quantifying the lack/presence of structure in a signal called, ``deviation from stochasticity" (DS) measure to characterize resting-state fMRI time series. The hypothesis is that differences in the level of structure in the time series can lead to discrimination between the subject groups. In this work, an autoencoder-based model is utilized to learn efficient representations of data by training the network to reconstruct its input data. The proposed methodology is applied on fMRI time series of 50 healthy individuals and 50 subjects with Alzheimer's Disease (AD), obtained from publicly available ADNI database. DS measure for healthy fMRI as expected turns out to be different compared to that of AD. Peak classification accuracy of 95\% was obtained using Gradient Boosting classifier, using the DS measure applied on 100 subjects.     

\end{abstract}


\section{Introduction}

Approximately 55 million people worldwide \cite{b14} suffer from dementia, which is the general term for global cognitive decline, leading to functional impairment. Alzheimer's disease (AD) is the most prevalent kind of dementia, accounting for 60–70\% of cases \cite{b22}, affecting the patient's health and quality of life. The formation of intracellular neurofibrillary tangles (NFTs) and amyloid-beta plaques are the primary causes of the pathogenesis of AD \cite{b22}, which ultimately leads to neuronal death. Because of its complex pathogenesis, it is challenging to accurately diagnose and classify AD. Also, the diagnosis and therapy lack a conclusive or broadly applicable method \cite{b23}. So, it is crucial for clinical research to create an efficient diagnostic technique for AD classification.  

Functional Magnetic Resonance Imaging (fMRI) is a non-invasive and non-ionizing technique used for acquiring the images from the brain and analyzing the functional properties and connectivity from the Blood Oxygenation Level Dependent (BOLD) signal. By measuring changes in blood flow and oxygenation levels associated with neural activity \cite{b1}, fMRI helps map brain regions involved in specific tasks or cognitive processes with high spatial resolution. This capability allows for the investigation of how different brain areas interact and specialize in performing various functions, from sensory perception and motor control to higher-order cognitive functions such as language processing and decision-making \cite{b3}. 

Furthermore, fMRI aids in the localization of brain abnormalities and provides insights into the mechanisms underlying neurological and psychiatric disorders \cite{b4}. Its ability to capture dynamic changes in brain activity in real-time has positioned fMRI as a cornerstone tool in neuroscience, advancing our understanding of the human brain's complexities and paving the way for innovative approaches in both basic and clinical neuroscience research \cite{b4}. 

Resting state fMRI (rs-fMRI) measures spontaneous neural activity in the brain while a subject is at rest, not performing any specific tasks or stimuli. It is an effective way to investigate the brain's extensive functional integration. A resting-state BOLD signal contains enough function-related information to identify meaningful networks and merits further study to gain a deeper understanding of the brain function \cite{b5}. Moreover, voxel-wise differences can be observed on the brain regions of healthy and unhealthy subjects with the help of rs-fMRI analysis \cite{b6}. Hence, it is widely used for studying Mild Cognitive Impairment (MCI) and AD in patients \cite{b23}.  

An fMRI dataset is made up of a concatenated string of volumes, also referred to as a ``run" or a ``scan" \cite{b7}. It is a collection of voxels, each of which has a time series with an equal number of time points as volumes collected in a given session. If we can identify the voxels whose time-course correlates with a known pattern through experimentation, then we can confirm which of the brain regions are coordinated together to perform a particular activity \cite{b7}. By analyzing the areas with an increase or decrease in BOLD signal over time, we can infer which regions are involved in different cognitive processes \cite{b8}, such as memory, language, or motor function. Studying the time series data provides insights into the temporal dynamics of brain activity \cite{b9} and are useful for mapping out regions of the brain that are active during specific tasks or in resting states. It helps to understand how neural networks interact over time, both within specific brain regions and across distributed networks. 

Changes in rs-fMRI time series patterns can potentially serve as biomarkers for neurological and psychiatric disorders such as Alzheimer's disease (AD) \cite{b10, b11}, schizophrenia, or depression \cite{b12}. Examining fMRI time series across different individuals allows for the exploration of variability in brain function. This variability can shed light on individual differences in cognitive abilities, preferences, and susceptibilities to neurological disorders \cite{b13}. 

fMRI has the ability to detect neuronal activation in the brain accurately and it is influenced by the signal-to-noise ratio (SNR) in the time-series data. The important factors to be considered while analyzing the time-series data include: 

\begin{itemize}
    \item Signal in fMRI: The signal in fMRI refers to the changes in blood flow and oxygenation levels that occur in response to neuronal activity. When neurons in a specific brain region become active, they require more oxygen. This leads to an increase in blood flow to that region, resulting in a detectable change in the MRI signal. 

    \item Noise in fMRI: Noise refers to any unwanted variation or interference in the MRI signal that is not related to neuronal activity. Sources of noise in fMRI include physiological noise (e.g., heartbeat, breathing), motion artifacts (e.g., head movement), and scanner-related noise (e.g., electromagnetic interference).

    \item Signal-to-Noise Ratio (SNR): SNR is a measure that quantifies the strength of the signal relative to the level of background noise in the data. A high SNR means that the signal (neuronal activity-related changes) is much stronger than the noise, making it easier to distinguish true neuronal activation from random fluctuations.

    \item Sensitivity of fMRI: The sensitivity of fMRI refers to its ability to accurately detect and localize neuronal activation. It depends critically on the SNR of the acquired fMRI data. Higher SNR enhances sensitivity because it reduces the likelihood of false positives (detecting activation where there is none) and improves the ability to detect small, genuine changes in neuronal activity.

    \item Impact of SNR on Data Quality: Low SNR can reduce the reliability and interpretability of fMRI results. It may obscure true neuronal activation, leading to weaker statistical evidence or false conclusions about brain function. Therefore, efforts in fMRI research focus on improving SNR through advanced imaging techniques, noise reduction methods, and careful experimental design.
\end{itemize}

Hence, the sensitivity of fMRI in detecting neuronal activation is dependent on the relative levels of signal and noise in the time-series data \cite{b2}. ``Intrinsic brain noise is dynamic since it is involved in brain activity over time", says, Scarciglia et al. (2024). It is necessary to comprehend the underlying physiology and biophysics in order to distinguish between a component's signal and noise \cite{b18}.   

This aids to the consideration of a factor called ``Stochasticity" in fMRI time series. It refers to the presence of random variability and unpredictability in the measurements of the BOLD signal over time \cite{b17}. This variability can arise from various sources and can impact the interpretation and analysis of fMRI data. It is characterized by the following key elements:

\begin{itemize}
    \item Random Variables: The values of the time series at each time point are random and subject to probabilistic variability. This randomness can stem from inherent variability in the underlying processes being observed or from external sources of randomness.

    \item Indexing in Time: The sequence of observations is indexed according to time, meaning that each observation is associated with a specific moment or interval in time.
\end{itemize}

The sources of stochasticity may include: physiological variability (cardiac and respiratory effects), scanner-related variability (the noise generated by the MRI scanners), subject-related variability (head motion and vascular differences), environmental and experimental factors (ambient noise and experimental design) and inherent signal properties (where the BOLD signal itself exhibits stochastic characteristics due to its physiological basis) \cite{b17}. 

Stochasticity in fMRI time series data can impact the reliability and interpretability of findings in several ways:

\begin{itemize}
    \item Reduced Signal-to-Noise Ratio (SNR): Higher stochastic variability increases the difficulty of separating true signal (neural activity-related changes) from noise (random fluctuations), leading to decreased SNR.

    \item False Positives and Negatives: Random fluctuations can mimic or mask true neural activity patterns, potentially leading to false positive or negative results in analyses.

    \item Reproducibility Issues: Variability introduced by stochasticity can affect the reproducibility of findings across different studies or experimental conditions.
\end{itemize}

Stochastic time series are typically analyzed using probabilistic and statistical methods. This includes examining the distribution of values, calculating measures of central tendency and variability, and modeling the temporal dependence structure (autocorrelation) of the time series \cite{b7}. 

Studies are going on to detect AD, in order to find solutions and provide the treatment to patients suffering from the disease \cite{b20} \cite{b21}. There is no permanent cure for it, but we can prevent it from further damage. Several networks of the brain when analyzed can show the areas being activated and deactivated at a point. With the help of fMRI and software-based processing of the acquired images, we are able to visualize different brain regions and statistically analyze the structural and functional aspects. The default mode network (DMN) can be used to figure out the regions which are correlated to each other and those which overlap with numerous regions that are significantly impacted by external variations \cite{b19}. Regression of these variables helps to improve the fMRI analysis and detect correlated regions at rest correctly. 

Here, we are proposing an application of ``Deviation from Stochasticity" measure on fMRI time series of AD and HC subjects, that will act as a biomarker for the classification of AD-affected individuals and healthy controls. This method has been developed by Pradeep et al. (2023) and tested on time series of black hole data and has given sufficient results to detect noise in the data. If we can measure the deviation from stochasticity in fMRI time series from these ROIs of the DMN, then we can detect the regions in the brain which are showing significant changes or varying patterns and contributing towards disease classification among individuals.

\section{Related Works}

Studies have discovered that AD exhibits anomalous default mode network (DMN) activity. Chouliaras et al.(2023) points out that the studies have showed changes in resting state functional connectivity and have identified unique networks and regions impacted in each dementia \cite{b14}. Dara et al. (2023) describes different ways to diagnose AD. Machine learning models like Support Vector Machine (SVM) and Convolutional Neural Network (CNN). The existing challenges involved in proper classification of AD are inadequate data samples, absence of intelligent feature selection techniques and the need for methods to rectify the error rates \cite{b14} during classification.

Alzheimer’s disease can be diagnosed using deep learning and machine learning techniques \cite{b23}. Several research works have been conducted to classify the individuals into healthy and AD-affected ones. Dhakal et al. (2023) have utilized machine learning models like support vector machine (SVM), random forest classifier, logistic regression model, etc., for classifying the subjects based on the fMRI images \cite{b20}. 

Zhang et al. (2019) in their work, computed the discrete probability distribution of the co-activity of distinct brain areas in multi-scale time series data at different intervals. While examining the causation and correlation between the fMRI data, the contextual information was considered. To quantify the similarity between co-activity intensities of two objects of brain functional connectivity, they developed a novel technique based on time-series. Then they applied SVM on time-series features, for disease classification \cite{b31} and have achieved an accuracy of 0.8935\%. 

Khazaee et al. (2017) performed multivariate Granger Causality analysis on rs-fMRI on healthy controls and patients with AD and MCI. The graph measures obtained as output were fed into a machine learning model. To choose the best subset of features, filter and wrapper feature selection techniques were used on the original feature set. Using the best features and the naïve Bayes classifier, a 93.3\% accuracy rate for the classification of AD, MCI, and HC was attained \cite{b23}. They also pointed out the limitation in achieving a high performance in segregating the three groups.   

Lund et al. (2006) have suggested an alternate method to the data-driven calculation of serial correlation in fMRI. The residual autocorrelation is frequently interpreted as a proof of unmodeled, or known, sources of variance. They modelled several factors (that induce autocorrelation within the design matrix) using their ``nuisance variable regression" (NVR) approach \cite{b15}. These factors include residual movement effects, hardware-related low-frequency drift, and aliased physiological noise. The method was successful in capturing the hypothesized noise sources, which include respiratory and cardiac effects (Razavi et al., 2003). In fact, when handling serial correlation, the NVR technique outperformed other approaches like high-pass filtering. Though this strategy may have some appeal, it still depends a great deal on the noise sources being accurately defined, and data cleaning will still be necessary to account for additional unmodeled sources of temporal correlation \cite{b16}.  

Wu et al. (2021) proposed a novel method for AD diagnosis using Sample Entropy to measure the neural complexity of the brain causality network. To calculate the brain's causality series, rs-fMRI data from 29 AD patients and 30 cognitive normal (CN) controls were subjected to Granger Causality analysis using a sliding temporal window. Using the agglomerative hierarchical clustering algorithm, they clustered these causality series, and the sample entropy of the clusters was calculated to serve as the classification features. They used classifiers like, XGBoost, SVM cluster, Random Forest, and SVM. Using the best feature subsets and the SVM classifier, an accuracy of 89.83\%, a sensitivity of 90.00\%, and a specificity of 89.66\% were attained \cite{b33}.

Boaretto et al. (2021) designed a method using permutation entropy, to classify stochastic and chaotic signals and find out the strength of correlations. First, they took time series data of different signals. Then they generated a time series of flicker noise. An Artificial Neural Network (ANN) was trained to predict the parameter in noise. Frequency of occurrence of different patterns determine the probability for calculating permutation entropy (PE). The PE of time series and that of noise are compared and hence classified into stochastic and chaotic signals \cite{b25}. 

Another way to understand the stochastic dynamics is the non-parametric approach used by Anvari et al. (2016). To distinguish between diffusive and jumpy stochastic behaviors, as well as deterministic drift factors, they have used a stochastic dynamical jump-diffusion modeling method. They demonstrated that all of the unknown functions and coefficients of this modeling can be obtained directly from time series measurements. Through a data-driven inference of the deterministic drift term and the diffusive and jumpy behavior in brain dynamics from multi-channel electroencephalographic recordings of ten epileptic patients, they showed that dynamics may be described as a stochastic process with a smaller mean diffusion coefficient and mean jump amplitude \cite{b28}. 

The dynamic changes of functional characteristics obtained from regional mean time series of rs-fMRI were modelled by Suk et al. (2016). They developed a Deep Auto-Encoder (DAE) to find the non-linear functional links among brain regions. After obtaining the features, they employed a Hidden Markov Model (HMM) to calculate the dynamic properties of the internal states of the functional networks in rs-fMRI. They constructed a generative model with an HMM and estimated the chance of the input features of rs-fMRI belonging to the group, i.e., MCI or HC. Using graph theory, they examined the functional connectivities. They obtained an accuracy of 72.58\% (for ADNI2 dataset) \cite{b27}. 

To study the effective functional connectivity among brain regions, Dynamic causal modelling (DCM) was used widely \cite{b29}. They have assessed stochastic DCM in relation to deterministic variants. Both DCM variants can account for neuronal fluctuations or noise, according to Monte-Carlo simulations. Signal-to-noise ratios and non-linearities in the neural evolution function determine their relative effectiveness in terms of network identification. They also concluded that the estimation accuracy is probably affected by the fMRI sample rate as well as the length of the entire session.

Even though many research works have been carried out to classify the AD-affected patients from healthy controls (HC) using structural MR images \cite{b30} \cite{b34} \cite{b35}, there isn't a widely applicable technique that can be utilized to preprocess, analyze and categorize the fMRI time series into different classes based on the varying characteristics and stochastic behavior at different time points. Recently. Pradeep et al. (2023) has devised a novel method that applies multi-scale reconstruction techniques and the prominence of dissimilarity curve peaks on data in the time and frequency domains. Using an auto-encoder, the ``deviation from stochasticity" \cite{b36} in black-hole time series was measured. The DS measure was found out to be small for a stochastic signal, whose behavior is constant throughout a range of time scales; whereas, for a non-stochastic signal, this measure turned out to be comparatively higher. This is tested on synthetic data as proof of concept, and the final conclusion was that, ``the DS values less than 1.5 indicate stochastic signals, while those greater indicates non-stochastic signal" \cite{b36}. 

In our study, we are applying the concept of quantifying deviation from stochasticity (DS) in rs-fMRI time series, to check whether a uniform structure is present in the signal or not. The signal comprises of the actual data along with some amount of noise, making it difficult for classification into healthy and unhealthy people. With the DS measure, we will be able to discriminate the stochastic behavior in the time series of the brain in both healthy and AD-affected individuals. We are aiming to classify subjects based on the DS measure as well as identify the significant regions of interest that contribute towards the classification of the disease.

\section{Proposed Method}

\begin{figure}[htbp]
\centerline{\includegraphics{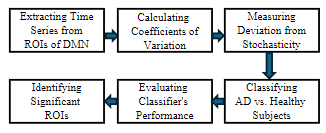}}
\caption{Proposed Methodology}
\label{fig:fig1}
\end{figure}

Deviation from Stochasticity (DS) is one of the techniques we can use to identify the complexities and disparities in the time series data for different brain regions. In our work, we describe the process in which fMRI time series is extracted from the ROIs of the DMN, the method of calculation of deviation from stochastic behavior for each subject corresponding to each ROI, and identification of significant features for classification. We are adopting the method used by the authors in \cite{b36}, which was applied for black hole data. Noise is present in fMRI time series data. If we identify the subjects with more stochastic signals from the brain regions, then we can easily classify them into healthy and AD classes. Our aim is to check whether the DS metric can accurately classify the subjects into 2 classes or not. Figure \ref{fig:fig1} represents the architecture of our proposed design. 

We are classifying the subjects into AD and healthy control groups based on a complexity measure in fMRI time series. The deviation from stochasticity (DS) is the measure that identifies how much the time series deviate from noise. If we find out this measure for each region of interest in the brain, we can see how different the pattern in the affected region is, compared to the healthy regions. 

The contributions of our paper include:
\begin{itemize}
    \item classification of subjects into healthy and AD groups using a novel method called, ``deviation from stochasticity" measure of DMN time series, where we are profiling subjects based on DS metric with respect to 34 different ROIs accurately,   
    \item focusing on DMN time series for identifying brain regions that contribute towards AD. 
\end{itemize}

\subsection{Loading of fMRI Data}

The dataset used for our study include resting state functional magnetic resonance images (rs-fMRI) of 100 subjects from Alzheimer's Disease Neuroimaging Initiative (ADNI) database. We are analyzing the brain activity from different regions when the person is in resting state, in order to determine the functioning of each region and their correlation with each other.  

fMRI data of 100 individuals are taken, out of which, 50 are healthy controls and 50 are affected by AD. The data is available in Digital Imaging and Communications in Medicine (DICOM) format. We upload it into the code and perform further analysis. DICOM images are converted into Neuroimaging Informatics Technology Initiative (NIFTI) images for processing. This conversion process begins by reading DICOM files and extracting voxel data arrays from multiple slices. Subsequently, these arrays are organized into a single 3D or 4D NIfTI file, integrating spatial and temporal information.

\subsection{Preprocessing of fMRI Data}

After conversion to NIfTI format, preprocessing steps are applied to the fMRI data to improve data quality and prepare it for subsequent analysis. The FSL (FMRIB (FMRI for the Brain) Software Library) toolbox is utilized for this purpose. These preprocessing steps may include:

\begin{itemize}  
    \item Slice Timing Correction: Adjusting for differences in acquisition time between slices within each volume to account for temporal offsets due to the sequential nature of slice acquisition.

    \item Motion Correction: Correcting for subject motion during the scan by aligning each volume to a reference volume or by estimating and correcting motion parameters.

    \item Artifact Removal: Addressing additional artifacts such as physiological noise (e.g., cardiac and respiratory fluctuations) using techniques like physiological noise regression.

    \item Intensity Normalization: Adjusting voxel intensities to correct for variations in signal intensity that can occur due to scanner differences or subject-specific factors.

    \item Smoothing: Applying spatial smoothing to the data to improve the signal-to-noise ratio and facilitate statistical analysis.

\end{itemize}

\subsection{Registering fMRI to MNI Template}

Next, we are performing spatial normalization to the data. To align individual brain images with a standardized anatomical space, we are registering fMRI data to the Montreal Neurological Institute (MNI) template. This process involves several steps to achieve spatial normalization. Initially, the fMRI data, typically acquired in subject-specific anatomical coordinates, undergo preprocessing to correct for distortions and ensure uniformity across the dataset. The individual anatomical images are coregistered to the high-resolution structural image or the MNI template. Subsequently, nonlinear transformations are applied to map each subject's brain into the MNI template space.

\subsection{Plotting the DMN with Several ROIs}

A parcellation scheme that covers various cortical and sub-cortical areas of the brain, the Dosenbach Atlas, derived from resting-state fMRI data, is used to divide the brain into 160 regions of interest (ROIs) based on functional connectivity patterns. The ``default mode network" (DMN) refers to the parts of the brain that are more active when at rest. It represents brain activity when not engaged in a particular cognitive task and the brain areas within this network are connected at rest \cite{b19}.

Here, the DMN is plotted with its ROIs. A binary mask is applied to the preprocessed fMRI dataset. This step selects voxels that are within the ROI mask. Each ROI represents a functionally coherent area, to map and analyze the brain network. This approach illustrates the synchronized activity among key brain areas during rest and provides insights into brain network organization and dynamics, aiding in understanding normal brain function and its alterations in neurological disorders.

\subsection{Extracting Time Series from the ROIs} 

Once parcellation is done, each of the ROIs within the DMN is identified by extracting the average time series of activity from the corresponding voxels within these regions across the fMRI dataset. For each voxel in the ROI mask, we retain the corresponding BOLD signal intensity values from the fMRI data over time. These time series represent the average neural activity within each ROI over time, capturing fluctuations in the BOLD signal that correlate with neuronal activity. Then, we calculate the time points based on the acquisition parameters. By iterating over the number of time points, we calculate each time stamp sequentially after ensuring that the time points align correctly with the extracted time series data for each ROI. Here, we are considering 34 ROIs of the DMN for the study.

We observed some variations in time series patterns between the paired ROIs. We are able to distinguish the subjects based on differences in DS values between 2 hemispheres of the brain. In an AD-affected person, the time series pattern of vmPFC 7 is significantly different from that of vmPFC 1. The amplitude differs in both subjects. This gives an insight into the differentiating characteristics of the brain.

\subsection{Detecting Stochasticity in Signals} 

Identifying stochasticity in fMRI time series involves recognizing patterns of variability that cannot be attributed solely to the neural processes of interest but rather arise from random or unpredictable sources. Here are a few methods we used to determine the stochasticity in fMRI data:

\begin{itemize}
    \item Statistical Analysis:
    \begin{itemize}
        \item Temporal Variability: Statistical measures such as variance, standard deviation, and autocorrelation coefficients are used to assess the degree of variability in the fMRI time series. Random fluctuations not related to neural activity typically exhibit irregular patterns and lack consistent trends over time.
    
        \item Noise Characteristics: Analysis of the frequency spectrum of the time series can reveal characteristic noise profiles. Stochastic noise tends to be broadband and not concentrated at specific frequencies, distinguishing it from task-related or physiological signals which may exhibit more structured spectral characteristics.
    \end{itemize}
    
    \item Control and Resting State Conditions:
    \begin{itemize}
        \item  Comparison with Baseline: During resting state fMRI scans or control conditions where no specific task is performed, the observed variability in the BOLD signal can be compared to periods of task engagement. Higher variability during baseline or rest periods may indicate stochastic influences.
    
        \item Null Models: Statistical models that simulate random fluctuations based on known noise sources (e.g., physiological noise models) can be used to test the hypothesis of stochasticity in the observed fMRI time series. Deviations from these null models suggest the presence of stochastic components.
    \end{itemize}
    
    \item Preprocessing of signal:
    \begin{itemize}
        \item Artifact Detection: Techniques such as motion correction algorithms can identify and mitigate artifacts caused by subject motion, which can introduce stochastic variability into the data.
    
        \item Signal-to-Noise Ratio (SNR): Evaluating the SNR of the fMRI time series provides insight into the relative strength of signal (neural activity) compared to noise (stochastic fluctuations). A lower SNR suggests higher stochastic influence.
    \end{itemize}

    \item Experimental Design Considerations:
    Replication Across Subjects and Studies: Consistency in stochastic patterns observed across different subjects or in replication studies strengthens the evidence for stochasticity. Inconsistencies may indicate specific sources of variability that can be identified and controlled for.
    
    \item Advanced Signal Processing Techniques:
    Various signal processing algorithms were utilized for distinguishing signal sources from stochastic noise components based on their statistical properties.

\end{itemize}

There are some features that characterize each signal. If the signal has no special structure, then:
\begin{itemize}
    \item Reconstructing it at different time scales leads to similar reconstructions. This is quantified as the ``Difference in Reconstruction" (DR).

    \item The rate of fluctuations seen in the original time-series is uniform. This is captured as the "Density of Peaks measure" (DP). 
\end{itemize}

These 2 measures are computed for every node in the DMN of the brain. Each ROI has 2 measures associated with it: Time-Scale Invariance (TSI) and Density of Peaks Measure (DPM)

\subsection{Calculating Deviation from Stochasticity}

For computing the KL Divergence and performing further calculations, we are using the code repository available in ``https://github.com/csai-arc/blackhole\_stochasticity\_measure". 

We are adopting an autoencoder-based time-invariant representation for measuring deviation from stochasticity in fMRI time series (Pradeep et al., 2023). The time series signal is divided into a number of windows. We compute discrete Fourier transform (DFT) on each window to obtain spectral information. Further, we perform the following operations: (a) Crop the transformed window to a predefined length and (b) Compute the modulus of the transformed window.  

Features learned from consecutive windows can be called time-invariant if they are equal in the absence of change point (i.e. amplitude, mean, frequency should not change much within a window). We use the loss function to make the features time-invariant and also to have a good signal reconstruction. 

Now, we calculate the deviation from stochasticity (DS metric) in the fMRI time series. Let \( W_{k} \) be the window size such that \( W_{k} \) is in the range of \{5, 7, 9, ..., 50\}, and D be the dissimilarity curve of length T. We divide the dissimilarity curve into windows of size \( W_{k} \), achieving multi-scale resolution represented by:

\begin{equation}
d_{t}^{k} = \left[ D[t - W_{k} + 1], \ldots, D[t] \right]^{T}
\end{equation}

(i) We compute the window-wise bias from the input time-series signal (X) as shown: 
\begin{equation}
b_{t}^{k} = \text{Mean} \left[ X[t - W_{k} + 1], \ldots, X[t] \right]^{T}
\end{equation}

\begin{equation}
\text{where } b_{t}^{k} \in \mathbb{R}^{W_{k}}
\end{equation}

(ii) This bias is added to the dissimilarity curve (D) to obtain bias corrected dissimilarity curve for a specific window size as shown below:
\begin{equation}
\tilde{d}_{t}^{k} = [D[t - W_{k} + 1] + b_{t}^{k}, \ldots, D_{t} + b_{t}^{k}]^{T}]
\end{equation}

(iii) We compute KL divergence between input time-series signal and bias corrected dissimilarity curve as shown:

\begin{equation}
z^{k} = \text{KLdivergence}(X, \tilde{D}_{k})
\end{equation}

(iv) Coefficient of variation is computed as: 
\begin{equation}
CV_{1} = \left( \frac{\text{std}(z^k)}{\text{mean}(z^k)} \right) \times 100
\end{equation}

(v) We divide the prominence of peaks curve into windows of size Wk, achieving multi-scale resolution represented by:
\begin{equation}
p_{t}^{k} = \left[ P[t - W_{k} + 1], \ldots, P[t] \right]^{T}
\end{equation}

\begin{equation}
\text{where } p_{t}^{k} \in \mathbb{R}^{W_{k}}
\end{equation}

(vi) We compute coefficient of variation for each of the prominence of peaks curve windowed over length Wk and store it as an array as shown:
\begin{equation}
COV_{t}^{k} = \frac{\text{std}(P^{k})}{\text{mean}(P^{k})} \times 100 
\end{equation}

(vii) We calculate the coefficient of variation CV2 as:
\begin{equation}
CV_{2} = \frac{\text{std}(COV_{t}^{k})}{\text{mean}(COV_{t}^{k})} \times 100
\end{equation}

(viii) Then, we define “deviation from stochasticity” (DS)
as below:
\begin{equation}
DS = (CV_{1} * CV_{2})/100.
\end{equation}

Across multi-scale resolutions, \(CV_{1} \) captures the variation between the original and bias-corrected reconstruction.

\subsection{Comparison of DS Metric between Subjects and among ROIs}

Since we have both healthy and AD-affected individuals and their corresponding DS values, we are able to perform a comparison within subjects and within ROIs in 4 different ways: 

\begin{itemize}
    \item Variation in DS measure within healthy controls: While calculating the DS value for each healthy control, we observed some significant trends. 
    \item Variation in DS measure within AD-affected individuals: While calculating the DS value for each affected individual, we saw another pattern along the subjects.  
    \item Variation in DS measure between healthy and affected individuals for a fixed ROI: When we compared the DS values between healthy and affected people, we saw that the values were consistent among HC class and not consistent in AD class.
    \item Variation in DS measure among paired ROIs for each subject: We found significant changes in DS values between symmetrical ROIs, that is, the right and left sides of the brain were having different DS values for both HC and AD classes. This portrays an asymmetry between the regions, that helps identify the main regions deviating from stochasticity.  
\end{itemize}

\subsection{Ablation Study}

In our study, we first computed the values of coefficients of variation, CV1, CV2, for each of the 100 subjects. We were able to find significant differences between healthy and AD subjects. We also saw some changes within the healthy subjects and within the unhealthy subjects. Later, we fed the data into various classifiers. The SVM classifier gave an accuracy of 95\% and the GBC gave an accuracy of 85\% for the covariance-based classification of subjects. This shows that linear classifiers are able to classify subjects based on the covariance values more accurately than the sophisticated models like RFC and GBC. These results are convenient to show that the DS measure and the covariance measure are efficient for the accurate classification of subjects into AD and HC classes.

The figure \ref{fig:fig5} below shows the scatter plot that we obtained while analyzing the AD and HC subjects based on CV1 and CV2 values. It distinguishes the 2 classes as shown. Here, the x-axis represents CV1 values and y-axis represents CV2 values. Dataset 1 denotes healthy data and Dataset 2 denotes AD-affected data. 

\begin{figure}[htbp]
\centerline{\includegraphics{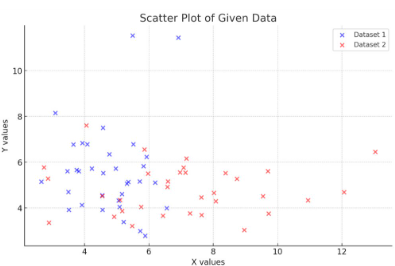}}
\caption{Scatter Plot separating AD and HC Subjects using CV1 and CV2 Values show discrimination between the 2 groups}
\label{fig:fig5}
\end{figure}

Then we applied SHapley Additive exPlanations (SHAP), that helps to explain the internals of the classifier trained on cognitive and clinical information, thus showing a possible link between diagnosis and patterns of feature relevancy \cite{b26}. It is used for explaining individual predictions made by machine learning models was applied by assigning a value to each feature.

The ROIs represented as features in the above graph include:
\begin{itemize}
    \item Feature 2: Anterior Prefrontal Cortex (aPFC) 5

    \item Feature 7: Inferior Temporal Cortex 72

    \item Feature 12: Occipital Cortex 136

    \item Feature 15: Post Cingulate Cortex 115

    \item Feature 16: Post Cingulate Cortex 111

    \item Feature 17: Post Cingulate Cortex 108

    \item Feature 18: Post Cingulate Cortex 93

    \item Feature 19: Post Cingulate Cortex 90 

    \item Feature 20: Post Cingulate Cortex 73

    \item Feature 24: Precuneus Cortex 94
\end{itemize}

\section{Results}

We applied the autoencoder-based deep learning model on the rs-fMRI time series data, to quantify the deviation from stochasticity for each subject. We obtained results for different subjects (healthy and AD classes) and for different ROIs. 

The DS measure was calculated for all the subjects corresponding to the 34 ROIs. These values were given as input to classifiers like, Logistic Regression (LR), SVM, Random Forest Classifier (RFC) and Gradient Boosting Classifier (GBC). We obtained a high accuracy of 95\% for both RFC and GBC. This implies that the linear classifiers are unable to classify the subjects accurately while the sophisticated models like RFC and GBC are capable of classifying subjects into healthy and AD groups based on the deviation form stochasticity measure. This proves that the DS values between healthy and unhealthy subjects can vary significantly. Hence, DS metric serves as a basis to classify subjects into healthy and AD groups. 

After the classification, we applied the t-Distributed Stochastic Neighbor Embedding (t-SNE) technique, that maps the high-dimensional data into a lower-dimensional space while preserving local relationships between points as much as possible, and obtained the plot as shown below in figure \ref{fig:fig7}:

\begin{figure}[htbp]
\centerline{\includegraphics{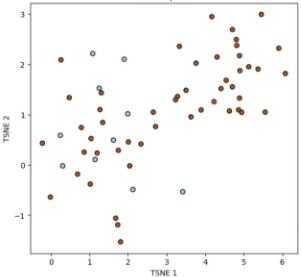}}
\caption{t-SNE Plot for Subjects based on DS Values clearly illustrates that 2 distinct clusters are formed, showing the goodness of the feature space}
\label{fig:fig7}
\end{figure}

Then we plotted the Receiver Operating Characteristic (ROC) and Area Under the Curve (AUC) in order to evaluate the performance of the classification models. The AUC for the RFC was 96\% and that for the GBC was 99\%. This indicates that the model has excellent discriminatory ability. Specifically, the GBC ranks 99\% of positive instances higher than negative instances on average.

\begin{figure}[htbp]
\centerline{\includegraphics{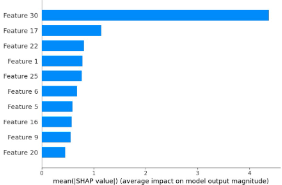}}
\caption{SHAP based on DS Values}
\label{fig:fig10}
\end{figure}

Then, SHAP technique is applied by assigning each feature an importance value for a particular prediction as shown in figure \ref{fig:fig10}. 

The ROIs represented as features in the above graph include:
\begin{itemize}
    \item Feature 1: Intraparietal Sulcus (IPS) 134 

    \item Feature 5: Fusiform Gyrus 84

    \item Feature 6: Inferior Temporal Cortex 91

    \item Feature 9: Medial Prefrontal Cortex (mPFC) 4

    \item Feature 16: Post Cingulate Cortex 111 

    \item Feature 17: Post Cingulate Cortex 108

    \item Feature 20: Post Cingulate Cortex 73 

    \item Feature 22: Precuneus Cortex 112  

    \item Feature 25: Precuneus Cortex 85

    \item Feature 30: Ventromedial Prefrontal Cortex (vmPFC) 1
\end{itemize}

After we confirmed the nodes which are contributing towards the AD vs. HC classification, we plotted the DS values among the HC and AD groups corresponding to the ROIs. We considered 4 different cases and obtained the plots accordingly. 

A consistent pattern in DS values in healthy controls and an abrupt pattern in the AD group was observed. This provides a good method to segregate the subjects based on the DS values.

\begin{figure}[htbp]
\centerline{\includegraphics{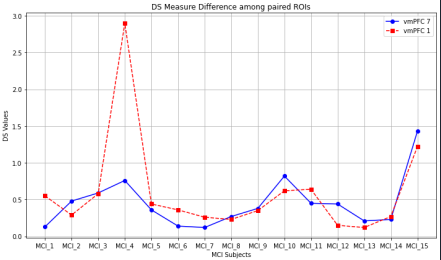}}
\caption{DS Measure Difference between vmPFC 7 \& 1 (right and left sides of brain) of AD Subject}
\label{fig:fig12}
\end{figure}

\begin{figure}[htbp]
\centerline{\includegraphics{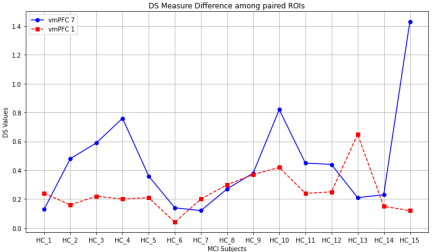}}
\caption{DS Measure Difference between vmPFC 7 \& 1 of HC Subject (right and left sides of brain)}
\label{fig:fig13}
\end{figure}

The figures \ref{fig:fig12} and \ref{fig:fig13} show the pattern of variation of DS measure in the ROIs which are paired in the brain like, vmPFC 7 and vmPFC 1. We can see how the pattern is different for the 2 regions. 

Then, we checked the pattern of DS measure for the Posterior Cingulate 108 and Posterior Cingulate 115 regions in 15 AD and 15 HC subjects. There was a variation in the pattern within 2 groups of subjects itself. This helps visualize the changes happening across the HC patients and those across the AD patients separately.

These studies investigate the disparities in the DS measure and prove that there exists a variation in the stochastic behavior of time series associated with 4 cases: within HC subjects, within AD subjects, among all subjects for a fixed ROI, and among the paired ROIs in the subjects.

\section{Discussion}

Alzheimer's disease (AD) is the most common cause of dementia globally and it is a chronic neurological illness with profound effects. In addition to being a medical problem, the condition involves a complex system that is yet poorly understood \cite{b37}. By analyzing the structural changes in the fMRI time series, we can identify the regions which are significantly deviating from the normal characteristics. In our study, we utilized the deviation from stochasticity measure to characterize the time series of healthy and AD-affected individuals. 

We are adapting the autoencoder-based model, utilized by Pradeep et al. (2023) to divide the fMRI time series into a number of windows, find the time-series invariance and the dissimilarity features. The deviation from stochasticity measure is calculated for each of the 34 ROIs of the DMN of the subjects after determining the coefficients of variation. This helps in categorizing the signals based on their stochastic behavior \cite{b36}. The DS measure in affected individuals were having a different pattern compared to that of healthy controls. The values were coming under different ranges for various ROIs as well for various subjects. This means that the fMRI time series contains signal along with noise, and this noise can be used to discriminate between signals in both groups of subjects.  

The use of machine learning models helped in classifying the subjects into healthy and AD-affected ones with a great accuracy. Zhang et al. (2019) applied SVM classifier on time-series features, for disease classification \cite{b31} and have obtained an accuracy of 0.8935\%. Using the best features and the naïve Bayes classifier \cite{b23}, a 93.3\% accuracy was obtained for Khazaee et al. (2017) while classifying AD, MCI and HC groups. Compared to these works, we have achieved a very high accuracy of 95\% using RFC and GBC for DS-based classification, making DS metric a biomarker for the AD Dementia vs. HC classification in subjects. So, our work incorporates a novel approach towards the classification of data that offers valuable results.

\section{Conclusion}
In this study, we are utilizing the deviation from stochasticity (DS) measure for identifying the disparities in time points of an fMRI data. With the DS metric, we are able to classify individuals into two classes: a) Group of individuals having AD and b) Group of healthy controls, based on the stochastic behavior of each region's time series. Clinicians can utilize this data to understand the characteristics of the time points across the entire fMRI data and finally confirm the regions which contribute towards the developing neuro-degenerative diseases. This will help in the study and monitoring of diseases at different regions of the brain and if applied earlier on unhealthy patients, can help reduce the risks associated with Alzheimer's disease. Hence, the proposed method can be widely used for classification purposes.

\section{Declaration of competing interest}

All authors declare no conflicts of interest.

\section{Acknowledgement}

We thank the Director, Dr. K. V. S. Hari, and the administration of the Centre for Brain Research, IISc, for the support provided throughout the study.

\vspace{12pt}

\end{document}